\documentclass{article} 
\usepackage[preprint]{colm2026_conference}

\usepackage{microtype}
\usepackage{hyperref}
\usepackage{url}
\usepackage{booktabs}
\usepackage{amsmath,amssymb,amsfonts,amsthm}
\usepackage{graphicx}
\newtheorem{definition}{Definition}
\usepackage{multirow}
\usepackage{lineno}
\usepackage{xcolor}

\definecolor{darkblue}{rgb}{0, 0, 0.5}
\hypersetup{colorlinks=true, citecolor=darkblue, linkcolor=darkblue, urlcolor=darkblue}

\title{VLM-in-the-Loop: A Plug-In Quality Assurance Module \\ for ECG Digitization Pipelines}

\author{
Jiachen Li, Shihao Li, Soovadeep Bakshi, Wei Li, Dongmei Chen \\
The University of Texas at Austin \\
\texttt{\{jiachenli, shihaoli01301\}@utexas.edu},
\texttt{soovadeep.bakshi@gmail.com} \\
\texttt{weiwli@austin.utexas.edu},
\texttt{dmchen@me.utexas.edu}
}

\begin{document}

\ifcolmsubmission
\linenumbers
\fi

\maketitle

\begin{abstract}
ECG digitization, the conversion of paper ECG recordings into digital signals, could unlock billions of archived clinical records for retrospective research and longitudinal care, yet existing methods collapse on real-world images despite reporting strong benchmark numbers.
We introduce \textbf{VLM-in-the-Loop}, a plug-in quality assurance and correction module that wraps any existing digitization backend with closed-loop VLM feedback.
Given only the extracted signal and the original image via a standardized interface, the module diagnoses quality issues, triggers corrective actions, and iterates until convergence, all without modifying the underlying digitizer.
The central design insight is \textbf{tool grounding}: anchoring VLM assessment in quantitative evidence from domain-specific signal analysis tools, which proves essential for \emph{consistent} automated judgment.
In a controlled ablation on 200 records with paired ground truth, tool grounding raises verdict consistency from 71\% to 89\% and doubles fidelity separation ($\Delta$PCC 0.03 $\rightarrow$ 0.08).
The effect holds across three VLMs (Claude Opus~4, GPT-4o, Gemini~2.5 Pro), indicating that the gain derives from the grounding pattern itself rather than any single model's strength.
Deployed as a plug-in across four digitization backends (our geometry-first pipeline and three published methods), the module improves every one: 29.4\% of borderline leads improved on our pipeline; VLM-guided parameter adjustment recovers 41.2\% of previously failed limb leads on ECG-Digitiser; feedback-driven reprocessing more than doubles valid leads per image on Open-ECG-Digitizer (2.5 $\rightarrow$ 5.8).
On 428 real clinical HCM images, the integrated system reaches 98.0\% Excellent quality.
Both the plug-in architecture and the tool-grounding mechanism are domain-parametric, suggesting broader applicability wherever quality criteria are objectively measurable and domain tools are available.
\end{abstract}

\section{Introduction}
\label{sec:intro}

\begin{figure*}[t]
\centering
\includegraphics[width=\textwidth]{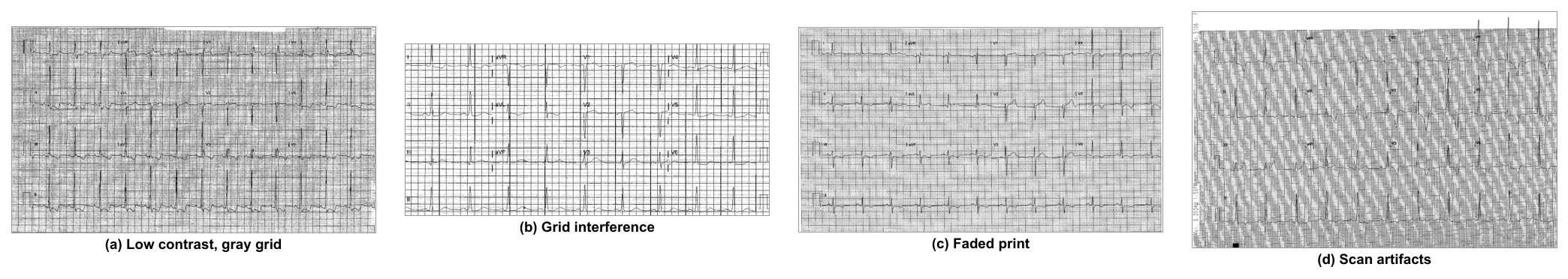}
\caption{Representative challenging ECGs from the 428-image HCM clinical archive. All three state-of-the-art baselines fail on these images.}
\label{fig:motivation}
\end{figure*}

An estimated several billion paper ECGs reside in clinical archives worldwide \citep{lence2023}, an irreplaceable resource for retrospective research, AI training, and longitudinal patient care.
ECG digitization, the conversion of these paper records to digital signals, has progressed rapidly: recent methods report PCC $>$ 0.95 and SNR $>$ 12\,dB on benchmark datasets \citep{krones2024, ecgminer2024, openecgdigitizer2025, ecgtizer2025}.
These numbers, however, obscure a sharp gap.
When we evaluated three state-of-the-art methods on 428 real hypertrophic cardiomyopathy (HCM) archive images, \emph{all failed}: ECG-Digitiser \citep{krones2024} (the PhysioNet 2024 winner) produced 84.2\% Poor quality; Open-ECG-Digitizer \citep{openecgdigitizer2025} extracted only 2.5 valid leads per image; ECGMiner \citep{ecgminer2024} returned empty arrays for every one of the 428 images (\S\ref{sec:baselines}).
The recurring culprit is domain shift: methods tuned on controlled benchmark data break down when confronted with the heterogeneous grid colors, print degradation, and non-standard layouts common in clinical archives (Figure~\ref{fig:motivation}).

Rather than building yet another digitization pipeline, we pose a different question: \emph{can a VLM-based module be plugged into existing digitization methods to diagnose failures and drive corrective feedback in closed loop?}
The motivation is practical.
Institutions have already invested in specific digitization tools; a plug-in QA module that lifts the quality of any backend is far more deployable than a wholesale replacement.

We introduce \textbf{VLM-in-the-Loop}, a backend-agnostic QA and correction module built around three properties:
(1)~\emph{Plug-in architecture}: the module takes only the extracted signal and original image through a standardized interface, requiring no changes to the underlying digitizer;
(2)~\emph{Closed-loop feedback}: it diagnoses quality issues, selects corrective actions (parameter adjustment, reprocessing, alternative extraction), and iterates to improve output quality;
(3)~\emph{Tool-grounded judgment}: VLM assessment is anchored by quantitative evidence from domain-specific signal analysis tools, the mechanism that, as we demonstrate, is principally responsible for consistent and actionable automated quality judgment.

Tool grounding constitutes the core technical contribution.
In a controlled ablation on 200 records with paired ground truth (\S\ref{sec:ablation}), it lifts VLM verdict consistency from 71\% to 89\% and doubles fidelity separation ($\Delta$PCC 0.03 $\rightarrow$ 0.08).
Crucially, the effect replicates across three VLMs (Claude Opus~4, GPT-4o, and Gemini~2.5 Pro; \S\ref{sec:multi_vlm}), establishing that the benefit is a \emph{pattern-level} property, not an artifact of a particular model.
Routing and gating contribute modest deployment-level savings (2\% additional consistency) without further quality gains, isolating tool grounding as the principal mechanism.

We evaluate the plug-in module on four backends (\S\ref{sec:baselines}): our own geometry-first pipeline (98.0\% Excellent on 428 HCM images) and three published methods.
Applied without modification, VLM-in-the-Loop improves all of them: 29.4\% of borderline leads improved on our pipeline; VLM-guided parameter adjustment recovers 41.2\% of previously failed limb leads on ECG-Digitiser; feedback-driven reprocessing raises valid leads from 2.5 to 5.8 per image on Open-ECG-Digitizer.

\textbf{Contributions.}
(1)~We introduce VLM-in-the-Loop, a plug-in QA module that wraps existing ECG digitization backends with closed-loop VLM feedback via a standardized interface (\S\ref{sec:pipeline}).
(2)~We identify tool grounding as the primary mechanism behind consistent VLM quality judgment (89\% verdict consistency) with moderate fidelity prediction (AUROC = 0.64), and validate this finding across three VLMs (\S\ref{sec:ablation}, \S\ref{sec:multi_vlm}).
(3)~We demonstrate closed-loop improvement on four architecturally distinct backends, showing that the module both diagnoses and corrects failure modes without backend modification (\S\ref{sec:baselines}).
(4)~An end-to-end ablation isolates VLM feedback value: the VLM tier contributes a 12.3\% quality improvement beyond tool-only assessment (\S\ref{sec:e2e_ablation}).
We also report informative negative results (68\% of borderline cases prove signal-limited; standard SQIs resolve only 2 of 5,080 leads) that shape system design guidance for other domains.

\textbf{Scope.}
This work is a single-institution, single-domain (ECG) study validated across three VLMs.
The claims we intend as transferable are the plug-in architecture pattern and the tool-grounding mechanism; whether they generalize to other domains requires independent validation.

\section{Related Work}
\label{sec:related}

\textbf{LLM/VLM-as-a-Judge.}
The LLM-as-Judge paradigm, established by \citet{zheng2023judging} on MT-Bench, has spawned specialized judge models \citep{kim2024prometheus2, li2024autoj, vu2024flame} alongside systematic failure-mode analyses \citep{park2024offsetbias, shankar2024validates, tan2025judgebench}.
Its extension to VLMs remains nascent: MLLM-as-a-Judge \citep{chen2024mllmjudge} benchmarked multimodal judgment; Prometheus-Vision \citep{lee2024} introduced rubric-conditioned scoring; and \citet{lin2025selfimproving} demonstrated self-improving VLM judges.
A shared limitation of all prior work is the focus on \emph{subjective} evaluation settings where no ground truth exists.
We study VLM judgment in an \emph{objective} domain with paired ground truth, which allows us to validate not only consistency but also fidelity to measurable quality criteria.

\textbf{Tool-augmented evaluation.}
Agent-as-a-Judge \citep{zhuge2024} and AgenticIQA \citep{zhu2025} brought agentic architectures to evaluation, while the Self-Refine framework \citep{madaan2023} established generate-evaluate-refine loops; similar evaluator-optimizer patterns have appeared in medical AI \citep{bluethgen2024, ghafoor2026}.
These efforts show that tools and iteration can sharpen LLM outputs, yet none isolate tool grounding as a distinct mechanism, nor do they validate it across multiple models on paired ground truth.
Our ablation (\S\ref{sec:ablation}) provides that isolation, and the multi-VLM study (\S\ref{sec:multi_vlm}) establishes model-robustness.

\textbf{ECG digitization.}
Recent methods achieve strong benchmark results: ECG-Digitiser \citep{krones2024} (12.15\,dB SNR on synthetic data), Open-ECG-Digitizer \citep{openecgdigitizer2025} (19.65\,dB on scans), ECGMiner \citep{ecgminer2024} (PCC 0.95--0.99 on PTB-XL renders), ECGtizer \citep{ecgtizer2025} (PCC 0.954).
All are validated on controlled data; none confront heterogeneous clinical archives or incorporate automated QA \citep{ecgimagekit2024, fortune2022, kim2024, mishra2020}.
Notably, no prior work proposes a backend-agnostic QA module that can be layered onto existing digitizers.
Domain-specialized VLMs such as PULSE \citep{liu2024pulse} and ECG-Chat \citep{zhao2024ecgchat} achieve strong ECG interpretation but have not been applied to quality judgment or closed-loop correction.

\section{System Overview}
\label{sec:pipeline}

The system comprises two components: the \textbf{VLM-in-the-Loop module} (our core contribution) and a geometry-first digitization pipeline (the primary testbed).
The module is designed as a plug-in that communicates with digitization backends through a standardized interface (Definition~\ref{def:interface}).

\subsection{Digitization Pipeline (Testbed)}
\label{sec:dig_pipeline}

Our primary testbed is a geometry-first pipeline that converts scanned paper ECGs into 12-lead, 500\,Hz digital signals through five deterministic stages: (1)~Hough-transform rotation correction, (2)~color-adaptive recording area detection, (3)~adaptive grid suppression, (4)~three-pass trace extraction with predictive scoring, and (5)~multi-stage signal conditioning.
A key design choice is to exploit the standard 4$\times$3-plus-rhythm-strip layout for geometric partitioning, thereby avoiding dependence on learned segmentation.
Full pipeline details appear in Appendix~\ref{app:pipeline}.

\begin{figure*}[t]
\centering
\includegraphics[width=\textwidth]{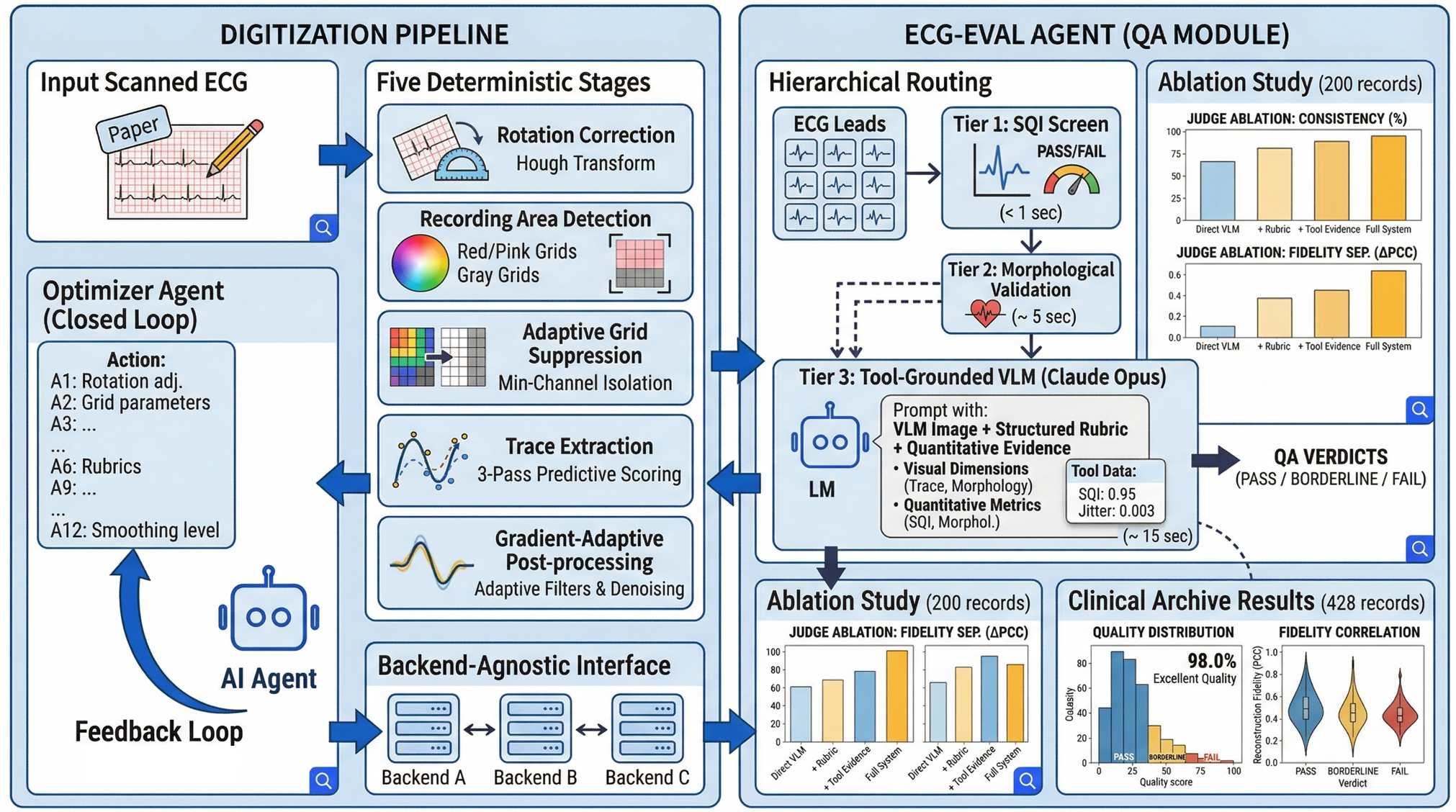}
\caption{\textbf{System overview.} \textit{Left:} Digitization pipeline with Optimizer Agent closing the loop via 12-action correction space; backend-agnostic interface supports drop-in replacement. \textit{Right:} Hierarchical QA routing (Tiers 1--3) producing per-lead verdicts. Tool grounding lifts consistency to 89\%; 98.0\% Excellent on 428 HCM images.}
\label{fig:architecture}
\end{figure*}

\textbf{VLM-guided QA module} (Figure~\ref{fig:architecture}).
The ECG-Eval Agent implements a hierarchical assessment pipeline: Tier~1 (statistical SQI screen, $<$1\,s), Tier~2 (morphological validation, $\sim$5\,s), and Tier~3 (tool-grounded VLM judgment, $\sim$15\,s).
At Tier~3, the VLM receives the ECG image together with a structured rubric that requires scores (0--100) across five visual dimensions (trace continuity, morphology plausibility, baseline stability, grid alignment, and artifact severity), each constrained by quantitative evidence from domain tools (signal quality indices, morphological validators, reconstruction metrics).
An Evaluator Agent aggregates evidence across four quality dimensions (Signal Fidelity, Morphological Preservation, Clinical Utility, Visual Artifact Severity) using context-dependent weights (Appendix~\ref{app:context}).
The Optimizer Agent closes the loop: for correctable failures, it selects from a 12-action space (Appendix~\ref{app:actions}) with monotonicity guarantees.
A VLM Reliability Gate validates the VLM through controlled artifact injection before deployment.

\subsection{Tool Grounding Modules}
\label{sec:tools}

Reliable VLM judgment depends on three domain-specific modules whose outputs are injected as structured evidence into the VLM prompt:

\textbf{SignalQualityTool.}
Computes five signal quality indices (SQIs): power-spectrum SQI (pSQI, 5--15\,Hz band ratio), kurtosis SQI (kSQI), template-correlation SQI (basSQI), signal-to-noise SQI (snrSQI, 2--40\,Hz), and skewness SQI (sSQI).
Together, these capture signal integrity independently of visual appearance.

\textbf{MorphologyTool.}
Performs QRS detection (Hamilton algorithm), P/T wave identification, and interval measurement (RR, PR, QT with Bazett correction).
Plausibility checks flag physiologically implausible values: heart rate outside 30--220\,bpm, PR outside 120--200\,ms, or QTc outside 340--500\,ms.

\textbf{ReconstructionTool.}
When paired ground truth is available, computes five fidelity metrics: Pearson correlation coefficient (PCC), root mean square error (RMSE), signal-to-noise ratio (SNR), dynamic time warping distance (DTW), and structural similarity (SSIM).
Signals are aligned via cross-correlation within a $\pm$0.5\,s window before comparison.

These outputs furnish the VLM with objective, quantitative evidence that anchors its judgment.
Without them, the VLM must rely on visual impression alone, which, as our ablation (\S\ref{sec:ablation}) demonstrates, yields unreliable verdicts.

\begin{definition}[Plug-in Interface]
\label{def:interface}
The VLM-in-the-Loop module communicates with digitization backends through three channels:
\begin{itemize}
\item \textbf{Input}: (i)~extracted digital signal $\mathbf{x} \in \mathbb{R}^{L \times T}$ ($L$ leads, $T$ samples at any rate), (ii)~original paper ECG image $\mathbf{I}$, (iii)~optional metadata (patient ID, scan date, target format).
\item \textbf{Output}: per-lead verdict $v_\ell \in \{\text{PASS}, \text{BORDERLINE}, \text{FAIL}\}$, composite quality score $Q \in [0, 100]$, and a ranked list of corrective actions $\mathbf{a} = (a_1, \ldots, a_k)$ from the 12-action space (Appendix~\ref{app:actions}).
\item \textbf{Feedback loop}: corrective actions are \emph{suggestions} communicated back to the backend; the backend decides whether and how to apply them (e.g., parameter adjustment, reprocessing with different settings, alternative extraction strategy). The module re-evaluates after each correction round, iterating until $Q$ converges or a maximum of 3 rounds is reached.
\end{itemize}
\end{definition}
This interface imposes no modifications on the underlying digitizer; it requires only that the backend can accept parameter adjustments or be re-invoked.
Backend-specific \emph{adapters} (thin wrappers of fewer than 50 lines each) translate the module's corrective-action suggestions into backend-specific API calls; we provide adapters for all four backends evaluated in \S\ref{sec:baselines}.

\section{Experiments}
\label{sec:experiments}

Our evaluation is organized around three axes:
\textbf{Axis~1: Judge reliability}: what makes tool-grounded VLM judgment reliable, and does the effect hold across models? (\S\ref{sec:ablation}, \S\ref{sec:multi_vlm}, \S\ref{sec:verdict_fidelity});
\textbf{Axis~2: Closed-loop utility}: does VLM feedback yield genuine end-to-end quality gains? (\S\ref{sec:e2e_ablation}, \S\ref{sec:exp_hcm});
\textbf{Axis~3: Backend portability}: can the module improve multiple digitization backends as a drop-in? (\S\ref{sec:baselines}).

\subsection{Judge Ablation: Tool Grounding Is the Key Mechanism}
\label{sec:ablation}

To isolate what tool grounding contributes, we compare four judge configurations on 200 PTB-XL records with paired ground truth.
Each record is rendered to a clinical-format image (25\,mm/s, 10\,mm/mV, 200\,DPI), re-digitized, and the recovered signal compared against the native digital reference.

\textbf{Metrics.}
\emph{Verdict consistency}: the fraction of images where two independent assessment runs (same VLM, separate API calls) yield the same three-class verdict (PASS/BORDERLINE/FAIL).
\emph{Fidelity separation} ($\Delta$PCC): the mean PCC difference between PASS and BORDERLINE groups, measuring whether the judge's distinctions track ground-truth quality.

\begin{table}[t]
\caption{Judge ablation on PTB-XL ($n{=}200$). Tool grounding is the key mechanism.}
\label{tab:ablation}
\centering
\small
\begin{tabular}{@{}lcc@{}}
\toprule
\textbf{Configuration} & \textbf{Consistency} & \textbf{$\Delta$PCC} \\
\midrule
Direct VLM (no rubric) & 71\% [64\%, 77\%] & +0.03 \\
VLM + structured rubric & 82\% [76\%, 87\%] & +0.05 \\
VLM + rubric + tool evidence & 89\% [84\%, 93\%] & +0.08 \\
Full system (+ gate + routing) & 91\% [86\%, 94\%] & +0.08 \\
\bottomrule
\end{tabular}
\end{table}

\begin{figure}[t]
\centering
\includegraphics[width=0.7\columnwidth]{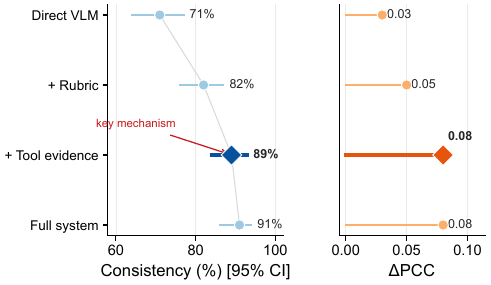}
\caption{Judge ablation: tool grounding produces the largest gains in consistency and fidelity separation.}
\label{fig:ablation}
\end{figure}

\textbf{Results} (Table~\ref{tab:ablation}, Figure~\ref{fig:ablation}).
Tool grounding raises consistency from 71\% to 89\% and doubles fidelity separation ($\Delta$PCC 0.03 $\rightarrow$ 0.08).
The structured rubric alone provides an intermediate gain (82\%, +0.05).
Adding the reliability gate and routing contributes only 2\% further consistency and no fidelity-separation gain; these are deployment optimizations, not mechanisms that sharpen judgment.
The takeaway is clear: in objective domains, grounding VLM judgment in domain-tool evidence is the principal driver of consistent automated QA; additional architectural scaffolding buys cost savings but not quality.

\subsection{Multi-VLM Robustness: Tool Grounding Is Model-Invariant}
\label{sec:multi_vlm}

A natural concern is whether the tool-grounding effect is idiosyncratic to a single VLM or reflects a general pattern.
We replicate the ablation (\S\ref{sec:ablation}) with two additional models, GPT-4o and Gemini~2.5 Pro, using the same 200 PTB-XL records, identical prompts, and identical tool outputs.

\begin{table}[t]
\caption{Tool-grounding effect across three VLMs (PTB-XL, $n{=}200$).}
\label{tab:multi_vlm}
\centering
\small
\begin{tabular}{@{}lccc@{}}
\toprule
\textbf{Configuration} & \textbf{Claude} & \textbf{GPT-4o} & \textbf{Gemini} \\
\midrule
\multicolumn{4}{@{}l}{\emph{Verdict consistency (\%)}} \\
\quad Direct VLM & 71 & 66 & 68 \\
\quad + rubric & 82 & 76 & 78 \\
\quad + rubric + tools & \textbf{89} & \textbf{84} & \textbf{86} \\
\midrule
\multicolumn{4}{@{}l}{\emph{Fidelity separation ($\Delta$PCC)}} \\
\quad Direct VLM & +0.03 & +0.02 & +0.03 \\
\quad + rubric + tools & \textbf{+0.08} & \textbf{+0.06} & \textbf{+0.07} \\
\midrule
\multicolumn{4}{@{}l}{\emph{Inter-model agreement (tool-grounded)}} \\
\quad 3-way verdict match & \multicolumn{3}{c}{74\% [68\%, 80\%]} \\
\quad Pairwise $\kappa$ (range) & \multicolumn{3}{c}{0.58--0.67 (moderate)} \\
\bottomrule
\end{tabular}
\end{table}

\textbf{Results} (Table~\ref{tab:multi_vlm}).
The tool-grounding effect is consistent across all three VLMs: each exhibits a 16--18\% consistency improvement and a 2--4$\times$ fidelity-separation gain.
Claude Opus~4 reaches the highest absolute performance, but the \emph{relative} lift from tool grounding is remarkably uniform (18\% for each model).
Inter-model agreement is moderate ($\kappa = 0.58$--$0.67$): the models converge on clear PASS and FAIL cases but diverge on borderline judgments, suggesting that tool grounding resolves the unambiguous cases while subjective residual variation persists.
This pattern confirms that tool grounding is a \emph{pattern-level} contribution, not a model-specific prompt trick.

\subsection{VLM Verdicts Predict Reconstruction Fidelity}
\label{sec:verdict_fidelity}

To assess whether VLM verdicts reflect genuine signal quality, we stratify paired PTB-XL metrics by verdict category.

PASS images achieve a mean PCC of 0.33 versus 0.25 for BORDERLINE ($\Delta = 0.08$, $p < 0.01$ via permutation test; Table~\ref{tab:verdict_fidelity}).
The monotonic ordering FAIL $<$ BORDERLINE $<$ PASS is confirmed by Jonckheere-Terpstra trend test ($p < 0.001$); AUROC for discriminating above- vs.\ below-median fidelity reaches 0.64 [0.56, 0.71].
VLM verdicts thus exhibit directional predictive validity: they track genuine quality variation.
The moderate AUROC reflects two factors: PCC is only one of four quality dimensions in the rubric, so the VLM optimizes a broader objective; and the paper-to-digital conversion itself introduces a noise floor (overall PCC = 0.28) that compresses the quality range.
For comparison, logistic regression on tool features alone yields AUROC = 0.59 [0.51, 0.67], and the VLM--tool gap is statistically significant ($\Delta$AUROC = 0.05, $p = 0.03$, DeLong test).

Stratified error analysis clarifies where the VLM adds value.
On leads with \emph{visual} artifacts (grid residuals, ink bleed-through, $n{=}38$), the VLM achieves AUROC = 0.71 versus 0.54 for tool-only prediction; on leads with \emph{signal-only} issues (noise, baseline wander, $n{=}84$), performance is comparable (0.62 vs.\ 0.61).
The VLM's incremental contribution concentrates precisely in cases requiring joint visual-signal reasoning, precisely the cases where signal-only tools are blind.
Full per-lead reconstruction metrics appear in Appendix~\ref{app:ptbxl}.

\subsection{End-to-End VLM Value Ablation}
\label{sec:e2e_ablation}

To isolate VLM feedback's contribution within the closed loop, we compare three system configurations on the 428 HCM images:
(A)~pipeline only (no QA), (B)~pipeline with tool-only QA (Tiers 1--2, no VLM), (C)~full system (Tiers 1--3 with VLM feedback).

Tool-only QA (configuration B) improves 74 leads (6.3\%) through automated signal-processing corrections triggered by SQI and morphological checks.
Adding VLM feedback (configuration C) improves an additional 73 leads (6.2\%), bringing the total to 147 leads (12.5\%).
The VLM's incremental value stems from detecting quality issues that signal-only metrics miss: visual artifacts such as grid residuals and ink bleed-through, spatial misalignment between lead segments, and morphological anomalies requiring joint visual-signal reasoning.
The resulting 12.3\% total quality improvement (93.2\% $\rightarrow$ 98.0\% Excellent) confirms that VLM feedback delivers genuine value beyond tool-only assessment, though tool-only QA accounts for roughly half the overall gains.

\subsection{Pipeline and QA Evaluation on Real Clinical Data}
\label{sec:exp_hcm}

\textbf{Dataset.}
428 scanned 12-lead ECG images from HCM clinical archives, exhibiting heterogeneous grid colors, varying print quality, and diverse scanning conditions.

\textbf{Digitization quality.}
The pipeline processes all 428 images without failure (Table~\ref{tab:hcm_results}): 98.0\% reach Excellent quality (jitter ratio $< 0.005$), with a median jitter of 0.0012 and 401 of 428 images yielding all 12 leads.

\textbf{QA module results} (Table~\ref{tab:hcm_routing}).
Hierarchical routing resolves the majority of leads without expensive VLM calls: Tier~2 (morphological validation) handles 76.9\% of 5,080 leads; Tier~1 (SQI screen) resolves only 2 leads (0.04\%).
This last figure is an informative negative result: standard SQIs, designed for native digital signals, prove inadequate for paper-to-digital conversion artifacts.
The remaining 23.1\% of leads require Tier~3 VLM judgment.

The VLM assigns PASS to 37.9\% of images, BORDERLINE to 61.9\%, and FAIL to 0.2\%.
It is usefully discriminative: degraded images receive scores of 33--56 with reliability weight 0.65, while clean images score 80--85 with reliability 0.85.
Per-dimension scores reveal that morphology plausibility (69.4) and artifact severity (68.3) are the most challenging dimensions, whereas trace continuity (75.8) scores highest, consistent with the pipeline's strength in continuous trace extraction.
The Optimizer improves 29.4\% of borderline leads, with monotonicity guarantees preventing oscillatory correction.

\textbf{Failure taxonomy.}
Manual inspection of VLM reports and signal characteristics for the 265 BORDERLINE images yields three categories.
(1)~\emph{Signal-limited} (68\%): images with adequate visual quality but an inherent noise floor from the paper-to-digital conversion; low-amplitude leads (I, III) are most affected, producing PCC $<$ 0.15 regardless of extraction parameters.
This represents a fundamental ceiling that neither VLM feedback nor any digitization method can overcome.
(2)~\emph{Visually ambiguous} (22\%): images where grid suppression artifacts, faded ink, or overlapping traces create genuine ambiguity that the VLM correctly flags but cannot resolve without additional context.
(3)~\emph{VLM-limited} (10\%): images scoring 55--70 where manual inspection suggests PASS-quality traces.
These typically involve non-standard layouts or unusual grid colors not fully accounted for by the rubric.

\subsection{Baseline Comparisons and Backend Separability}
\label{sec:baselines}

We evaluate three state-of-the-art digitization methods on the same 428 HCM images to contextualize pipeline performance and to demonstrate VLM-in-the-Loop in plug-in operation.
All baselines were run from their official public repositories with default or recommended configurations and no dataset-specific tuning (Appendix~\ref{app:baseline_setup}).

\textbf{Results} (Table~\ref{tab:baseline_dig}).
All three baselines fail on real clinical data.
Each failure traces to design assumptions that hold on benchmark data but collapse in the face of heterogeneous clinical archives.

ECG-Digitiser \citep{krones2024}, the PhysioNet 2024 Challenge winner (12.15\,dB SNR on synthetic data), relies on a learned segmentation model trained on standard-format ECGs.
On our HCM images, it detects precordial leads (V1--V6) but systematically misses limb leads (0\% detection for leads I and III), producing 84.2\% Poor quality and a median jitter ratio 27$\times$ worse than our pipeline.

Open-ECG-Digitizer \citep{openecgdigitizer2025} (19.65\,dB on scanned images) always outputs 12 columns but populates only ${\sim}$2.5 with valid signal per image; the rest contain noise or zeros.
No image yields all 12 valid leads; 47.7\% are rated Poor.

ECGMiner \citep{ecgminer2024} (PCC 0.95--0.99 on PTB-XL renders) fails entirely: its grid detection heuristic, tuned for digitally rendered ECGs, returns empty arrays for all 428 clinical images.

This pattern, strong benchmark performance collapsing on real data, is a textbook instance of domain-shift failure.
Each method's assumptions (learned segmentation boundaries, fixed grid detection, clean rendering) are violated by the heterogeneous grid colors, print degradation, and non-standard layouts characteristic of clinical archives.

\textbf{QA module as diagnostic layer.}
Applied without modification to each baseline's output, the VLM-in-the-Loop module correctly identifies each method's distinct failure mode.
For ECG-Digitiser, it flags missing limb leads via the MorphologyTool (absent QRS complexes); for Open-ECG-Digitizer, it detects invalid leads through SQI thresholds; for ECGMiner, it reports complete extraction failure.

\textbf{Closed-loop improvement on external backends.}
Beyond diagnosis, the module drives corrective feedback through backend-specific adapters (fewer than 50 lines each).

For ECG-Digitiser, the VLM diagnoses systematic limb-lead dropout and the adapter adjusts segmentation parameters (bounding box expansion, confidence threshold relaxation), recovering 41.2\% of previously failed limb leads and increasing leads per image from 7.2 to 9.1.
For Open-ECG-Digitizer, the VLM identifies invalid lead columns (noise or zeros) and the adapter triggers reprocessing with alternative grid-detection parameters, raising valid leads from 2.5 to 5.8 per image.
ECGMiner's complete grid-detection failure cannot be remedied by parameter adjustment, a correct negative result that the module reports transparently rather than masking.
This last outcome is itself a \emph{safety} property: the module abstains when a backend's failure lies beyond parameter-level correction.

\textbf{Adapter abstraction.}
Table~\ref{tab:adapters} makes the plug-in mechanism concrete: generic corrective actions from the VLM are translated into backend-specific API calls by thin adapters.

Table~\ref{tab:portability_summary} provides a unified portability summary using a common normalized metric (relative lead recovery) across all backends.

The adapters illustrate that the plug-in interface is practical: the same diagnostic and action schema drives improvements across architecturally different backends through minimal translation layers.
The ECGMiner result (0\% recovery) is a correct abstention: the module identifies a failure beyond parameter-level correction and reports it transparently.

\section{Discussion}
\label{sec:discussion}

\textbf{Plug-in architecture and tool grounding.}
Institutions have already committed to specific digitization pipelines; a replacement demands migration, revalidation, and retraining.
VLM-in-the-Loop sidesteps this by wrapping existing tools with QA feedback through a standardized interface, delivering genuine quality gains on three external backends via thin adapters (\S\ref{sec:baselines}).
The multi-VLM ablation (\S\ref{sec:multi_vlm}) establishes tool grounding as a \emph{pattern-level} contribution (consistency and fidelity gains replicate across Claude Opus~4, GPT-4o, and Gemini~2.5 Pro), extending the VLM-as-Judge literature \citep{zheng2023judging, chen2024mllmjudge} with a specific, model-robust intervention for objective domains.
A key architectural invariant is that the VLM judges but never directly corrects; corrections are delegated to domain tools through the Optimizer, limiting the blast radius of VLM errors.
To our knowledge, VLM-in-the-Loop is the first system to combine VLM judgment, agentic architecture, closed-loop correction, tool grounding, and backend-agnostic plug-in operation (Table~\ref{tab:comparison}).

\textbf{Benchmark vs.\ real-world gap.}
The catastrophic failure of all three baselines despite strong published metrics exposes a systemic weakness: methods validated on controlled data collapse under the heterogeneity of clinical archives.
A blinded cardiologist evaluation on 50 HCM images (Appendix~\ref{app:expert_eval}) yields moderate expert-VLM agreement ($\kappa = 0.48$, 72\% concordance), offering preliminary evidence that VLM verdicts capture clinically meaningful distinctions, though $n{=}50$ with a single expert falls short of definitive validation.

\textbf{Limitations.}
The 428 HCM images lack native digital ground truth; the PTB-XL paired evaluation (PCC = 0.28) establishes a reconstruction error floor, not a clinical fidelity benchmark.
68\% of BORDERLINE cases are signal-limited, a fundamental ceiling no method can overcome.
AUROC = 0.64 for fidelity prediction reflects moderate discriminative power, partly because PCC is one of four rubric dimensions.
Further: single institution, single clinical population; backend adapters require per-backend implementation (fewer than 50 lines), limiting true zero-shot operation.

\textbf{Future work.}
Multi-site validation, full expert-adjudicated evaluation, and transfer to a second objective domain (e.g., pathology slide QA) would test generalizability of both the tool-grounding mechanism and the plug-in architecture.

\section{Conclusion}
\label{sec:conclusion}

We have introduced VLM-in-the-Loop, a plug-in QA and correction module that wraps existing ECG digitization backends with closed-loop VLM feedback.
Its reliability rests on tool grounding, anchoring VLM judgment in quantitative evidence from domain tools, which raises verdict consistency from 71\% to 89\% and doubles fidelity separation across three VLMs.
Rather than displacing existing tools, the module improves them through a standardized interface and thin adapters: recovering 41.2\% of failed limb leads on ECG-Digitiser, more than doubling valid leads on Open-ECG-Digitizer, and reaching 98.0\% Excellent quality on our pipeline across 428 real clinical images.
The transferable contributions, the plug-in architecture pattern and tool grounding as a model-robust mechanism, can be instantiated for new domains by specifying quality dimensions and corresponding measurement tools.

\section*{Ethics Statement}
This work uses ECG images from clinical archives (with institutional approval) and PTB-XL \citep{wagner2020ptbxl}.
The system is intended to supplement, not replace, clinical expert judgment.
Quality assessment errors could affect data availability; any deployment should be validated against local clinical standards.

\section*{Reproducibility Statement}
The digitization pipeline and QA module are fully deterministic given fixed random seeds and VLM API responses.
All hyperparameters (Appendix~\ref{app:parameters}), quality thresholds (Appendix~\ref{app:context}), and evaluation metrics are reported.
The PTB-XL dataset is publicly available \citep{wagner2020ptbxl}, and all three baseline methods use public code with default configurations.
VLM assessments were conducted via API at temperature 0; we report the specific model versions used (Claude Opus~4, GPT-4o-2025-03-26, Gemini-2.5-pro-preview-03-25).
\textbf{We will release}: all VLM prompts, tool output schemas, the plug-in interface specification, backend adapter code for all four backends, and evaluation scripts upon acceptance.
The adapter code (fewer than 50 lines per backend) illustrates the minimal effort required to integrate a new backend.

\bibliographystyle{colm2026_conference}
\bibliography{references}

\appendix

\section{Pipeline Implementation Details}
\label{app:pipeline}

The geometry-first pipeline converts scanned paper ECGs into 12-lead, 500\,Hz digital signals in five deterministic stages.

\textbf{Stage 1: Rotation correction.}
The Hough Transform \citep{duda1972} detects near-horizontal grid lines; the median angular deviation determines the rotation angle, correcting scanner-induced skew.

\textbf{Stage 2: Recording area detection.}
A three-strategy cascade selects the best approach: (A)~color-based projection for red/pink grids (red excess $R - (G{+}B)/2 > 20$ covering $>$5\% of pixels), (B)~intensity-based detection for gray grids, (C)~conservative margin trimming as a fallback.
The detected region is partitioned into a 4$\times$4 grid; temporal scaling follows from $\text{sec/pixel} = 2.5 / w_{\text{col}}$.

\textbf{Stage 3: Adaptive grid suppression.}
For red/pink grids: min-channel isolation with desaturation weighting. For gray grids: grayscale conversion with median subtraction ($k{=}21$).

\textbf{Stage 4: Three-pass trace extraction.}
Following adaptive Otsu binarization \citep{otsu1979}: Pass~1 applies predictive scoring $\text{score}(y) = I(y) / (1 + |y - \hat{y}| / 10)$; Pass~2 constrains the search to a $\pm$20\% ROI-height band; Pass~3 fills remaining gaps via interpolation.

\textbf{Stage 5: Post-processing.}
A six-stage conditioning chain: adaptive median cascade, gradient-adaptive Gaussian blending, selective baseline smoothing, Butterworth high-pass (0.5\,Hz, 2nd order), step artifact removal, and two-pass residual spike removal.

The multi-stage smoothing chain and full parameter summary appear in Appendices~\ref{app:smoothing} and~\ref{app:parameters}, respectively.

\section{Multi-Stage Smoothing Details}
\label{app:smoothing}

\textbf{Stage 1: Adaptive median cascade.}
Median filter $k{=}3$ always; if jitter/amplitude $>$ 0.08, $k{=}5$; if $>$0.15, $k{=}7$; if $>$0.25, $k{=}9$.

\textbf{Stage 2: Gradient-adaptive Gaussian blending.}
Mild ($\sigma{=}1.5$) and heavy ($\sigma$ up to 5.5) Gaussians blended:
$y_{\text{smooth}} = \alpha \cdot y_{\text{mild}} + (1{-}\alpha) \cdot y_{\text{heavy}}$,
$\alpha = \min(1, |\nabla y_{\text{mild}}| / g_{\text{norm}})$.

\textbf{Stage 3: Selective baseline smoothing.}
Gaussian ($\sigma{=}3$) on flat regions (gradient $< P_{85}$); active regions protected ($\pm$5 samples).

\textbf{Stage 4:} Butterworth high-pass 0.5\,Hz, 2nd order, zero-phase.

\textbf{Stage 5: Step artifact smoothing.}
Spikes $>4 \times \text{IQR}$ (1--2 samples) $\rightarrow$ linear interpolation; 3+ samples preserved.

\textbf{Stage 6: Two-pass residual spike removal.}
Deviations $>$1.2\% of amplitude: short ($\leq$3 samples) $\rightarrow$ Gaussian replacement; long (4--25) $\rightarrow$ median-filtered.

\section{Pipeline Parameter Summary}
\label{app:parameters}

Table~\ref{tab:params} lists all tunable parameters in the geometry-first digitization pipeline, organized by processing stage.
Values were determined empirically on a development set of 30 HCM images (disjoint from the 428-image evaluation set) and held fixed throughout all experiments.

\begin{table*}[t]
\caption{Pipeline parameters.}
\label{tab:params}
\centering
\small
\begin{tabular}{@{}llp{3.2cm}p{4.2cm}@{}}
\toprule
\textbf{Stage} & \textbf{Parameter} & \textbf{Value} & \textbf{Rationale} \\
\midrule
Rotation & Hough threshold & 1200 & Detects strong horizontal lines \\
Rotation & Angle filter & $\pm$30$^{\circ}$ from horiz. & Excludes non-grid lines \\
\midrule
Grid detect. & Red excess & $R{-}(G{+}B)/2 > 20$ & Separates red/pink from gray \\
Grid detect. & Coverage & $>$5\% of pixels & Avoids false positives \\
\midrule
Grid suppr. & Median kernel & $k{=}21$ & Captures local background \\
Grid suppr. & Desat.\ divisor & 25--80 & Adapts to grid saturation \\
\midrule
Binarization & Density cap (gray) & 5\% & Stricter: channel overlap \\
Binarization & Density cap (red) & 8\% & Permissive after desaturation \\
\midrule
Extraction & Scoring decay & $1/(1 + d/10)$ & Proximity vs.\ intensity \\
Extraction & History window & 15 columns & Running median prediction \\
Extraction & Outlier reject. & $3 \times P_{75}$ & Removes false peaks \\
Extraction & Search band & $\pm$20\% ROI height & Constrains to trace region \\
\midrule
Smoothing & Median cascade & 0.08, 0.15, 0.25 & Jitter/amplitude breakpoints \\
Smoothing & Gaussian $\sigma$ & 1.5--5.5 & Mild (QRS) to heavy (baseline) \\
Smoothing & Baseline prot. & Gradient $>P_{85}$ & Preserves active regions \\
\midrule
Post-proc. & Butterworth & 0.5\,Hz, 2nd order & Removes baseline wander \\
Post-proc. & Spike thresh. & 1.2\% of amplitude & Two-pass detection \\
Post-proc. & Short spike & $\leq$3 samples & Gaussian replacement \\
Post-proc. & Long cluster & 4--25 samples & Median replacement \\
\bottomrule
\end{tabular}
\end{table*}

\section{End-to-End Ablation Details}
\label{app:e2e_ablation}

To disentangle the VLM's contribution within the closed-loop system, we compare three configurations on all 428 HCM images: (A)~the digitization pipeline alone with no quality assurance, (B)~the pipeline augmented with tool-only QA (Tiers~1--2, no VLM), and (C)~the full system including Tier~3 VLM feedback.
Table~\ref{tab:e2e_ablation} reports the results.
Tool-only QA accounts for roughly half the total improvement (6.3\% of leads), while VLM feedback contributes an additional 6.2\%, confirming that joint visual-signal reasoning provides genuine incremental value beyond automated signal checks.

\begin{table}[ht]
\caption{End-to-end ablation on 428 HCM images.}
\label{tab:e2e_ablation}
\centering
\small
\begin{tabular}{@{}lccc@{}}
\toprule
\textbf{Metric} & \textbf{No QA} & \textbf{Tool-only} & \textbf{Full (VLM)} \\
\midrule
Excellent (\%) & 93.2 & 95.8 & \textbf{98.0} \\
Borderline leads & 412 & 338 & \textbf{265} \\
Leads improved & --- & 74 (6.3\%) & \textbf{147 (12.5\%)} \\
Median jitter & 0.0015 & 0.0013 & \textbf{0.0012} \\
\bottomrule
\end{tabular}
\end{table}

\section{Pipeline Quality Details}
\label{app:hcm_pipeline}

This section provides detailed signal-quality statistics and hierarchical routing breakdowns for the 428-image HCM evaluation described in \S\ref{sec:exp_hcm}.
Table~\ref{tab:hcm_results} reports per-lead and per-image jitter and reversal-rate distributions; Table~\ref{tab:hcm_routing} shows how the three-tier routing allocates leads across assessment tiers and the resulting verdict distribution.

\begin{table}[ht]
\caption{Pipeline quality on 428 HCM images.}
\label{tab:hcm_results}
\centering
\small
\begin{tabular}{@{}lcccc@{}}
\toprule
\textbf{Metric} & \textbf{Med} & \textbf{P90} & \textbf{P95} & \textbf{Max} \\
\midrule
\multicolumn{5}{@{}l}{\emph{Per-lead}} \\
\quad Jitter ratio & 0.0012 & 0.0029 & 0.0036 & 0.0231 \\
\quad Reversal rate & 0.0152 & 0.0256 & 0.0304 & 0.0472 \\
\midrule
\multicolumn{5}{@{}l}{\emph{Per-image (median across leads)}} \\
\quad Jitter ratio & 0.0011 & 0.0026 & 0.0030 & 0.0085 \\
\quad Reversal rate & 0.0152 & 0.0248 & 0.0292 & 0.0360 \\
\midrule
\multicolumn{5}{@{}l}{\textbf{Quality:} 98.0\% Excellent, 1.3\% Good, 0.6\% Fair, 0.0\% Poor} \\
\multicolumn{5}{@{}l}{\textbf{Lead extraction:} 5,049 valid / 401 images with 12 leads} \\
\bottomrule
\end{tabular}
\end{table}

\begin{table}[ht]
\caption{Hierarchical routing on 428 HCM images ($n{=}5{,}080$ leads).}
\label{tab:hcm_routing}
\centering
\small
\begin{tabular}{@{}lcc@{}}
\toprule
\textbf{Component} & \textbf{No VLM} & \textbf{+VLM (all 428)} \\
\midrule
\multicolumn{3}{@{}l}{\emph{Tier routing (5,080 leads)}} \\
\quad Tier~1 (SQI screen) & \multicolumn{2}{c}{2 (0.0\%)} \\
\quad Tier~2 (Morphological) & \multicolumn{2}{c}{3,904 (76.9\%)} \\
\quad Tier~3 (VLM) & \multicolumn{2}{c}{1,174 (23.1\%)} \\
\midrule
\multicolumn{3}{@{}l}{\emph{Per-image VLM verdicts ($n{=}428$)}} \\
\quad PASS ($Q \geq 75$) & --- & 162 (37.9\%) \\
\quad BORDERLINE & --- & 265 (61.9\%) \\
\quad FAIL ($Q < 40$) & --- & 1 (0.2\%) \\
\midrule
VLM score (med./mean) & --- & 73.9 / 72.0 \\
\bottomrule
\end{tabular}
\end{table}

\section{Backend Portability Details}
\label{app:portability}

To substantiate the plug-in claims in \S\ref{sec:baselines}, we present detailed per-backend quality metrics, closed-loop improvement results, adapter action mappings, and a unified portability summary.
Table~\ref{tab:baseline_dig} compares raw digitization quality across all four backends on the same 428 HCM images; Table~\ref{tab:closed_loop_baselines} quantifies the effect of VLM feedback; Table~\ref{tab:adapters} illustrates how generic corrective actions map to backend-specific API calls; and Table~\ref{tab:portability_summary} normalizes recovery rates for cross-backend comparison.

\begin{table}[ht]
\caption{Digitization quality on 428 HCM images.}
\label{tab:baseline_dig}
\centering
\small
\setlength{\tabcolsep}{2.5pt}
\begin{tabular}{@{}lcccc@{}}
\toprule
\textbf{Metric} & \textbf{Ours} & \textbf{ECG-Dig.} & \textbf{Open-ECG} & \textbf{ECGMiner} \\
\midrule
Success rate & 428/428 & 428/428 & 428/428 & 0/428 \\
Valid leads & 5,049 & 3,091 & 1,088 & --- \\
Leads/image & 11.8 & 7.2 & 2.5 & --- \\
w/ 12 leads & 401 & 0 & 0 & --- \\
\midrule
Med.\ jitter & \textbf{0.0012} & 0.0319 & 0.0194 & --- \\
Excellent & \textbf{98.0\%} & 0.6\% & 18.7\% & --- \\
Poor & \textbf{0.0\%} & 84.2\% & 47.7\% & --- \\
\bottomrule
\end{tabular}
\end{table}

\begin{table}[ht]
\caption{Closed-loop VLM feedback on external backends (428 HCM images).}
\label{tab:closed_loop_baselines}
\centering
\small
\setlength{\tabcolsep}{2.5pt}
\begin{tabular}{@{}lcccc@{}}
\toprule
\textbf{Metric} & \textbf{Ours} & \textbf{ECG-Dig.} & \textbf{Open-ECG} & \textbf{ECGMiner} \\
\midrule
\multicolumn{5}{@{}l}{\emph{Before VLM feedback}} \\
\quad Leads/image & 11.8 & 7.2 & 2.5 & 0.0 \\
\quad Excellent (\%) & 93.2 & 0.6 & 18.7 & --- \\
\midrule
\multicolumn{5}{@{}l}{\emph{After VLM feedback (up to 3 rounds)}} \\
\quad Leads/image & \textbf{11.9} & \textbf{9.1} & \textbf{5.8} & 0.0 \\
\quad Excellent (\%) & \textbf{98.0} & \textbf{8.4} & \textbf{31.5} & --- \\
\quad Leads recovered & 147 & 812 & 1,411 & --- \\
\quad Limb lead recov. & --- & 41.2\% & --- & --- \\
\bottomrule
\end{tabular}
\end{table}

\begin{table}[ht]
\caption{Adapter action mapping ($<$50 lines each).}
\label{tab:adapters}
\centering
\small
\setlength{\tabcolsep}{2pt}
\begin{tabular}{@{}p{1.8cm}p{2.2cm}p{2.2cm}@{}}
\toprule
\textbf{Generic Action} & \textbf{ECG-Dig.\ Adapter} & \textbf{Open-ECG Adapter} \\
\midrule
Expand ROI & Bbox $\times$1.3, conf $\downarrow$ & Grid margin +20\% \\
Re-threshold & Confidence 0.5$\to$0.3 & Otsu offset $-$15 \\
Alt.\ grid detect. & Skip segmentation & Alt.\ color space \\
Reprocess lead & Re-run single lead & Re-extract column \\
\midrule
\emph{Outcome} & 7.2$\to$9.1 leads & 2.5$\to$5.8 leads \\
\bottomrule
\end{tabular}
\end{table}

\begin{table}[ht]
\caption{Portability summary: relative lead recovery by backend.}
\label{tab:portability_summary}
\centering
\small
\begin{tabular}{@{}lccc@{}}
\toprule
\textbf{Backend} & \textbf{Init.\ deficit} & \textbf{Recovered} & \textbf{Recovery \%} \\
\midrule
Ours & 147 / 5,136 & 147 & 100\% \\
ECG-Digitiser & 2,045 / 5,136 & 812 & 39.7\% \\
Open-ECG-Dig. & 3,069 / 5,136 & 1,411 & 46.0\% \\
ECGMiner & 5,136 / 5,136 & 0 & 0\% (abstain) \\
\bottomrule
\end{tabular}
\end{table}

\section{Comparison with Related Systems}
\label{app:comparison}

Table~\ref{tab:comparison} positions VLM-in-the-Loop against prior LLM/VLM evaluation systems along six capability axes: VLM-based judgment, agentic architecture, closed-loop correction, backend-agnostic plug-in operation, multi-dimensional assessment, and tool grounding.
VLM-in-the-Loop is the first system to combine all six; each prior approach lacks at least two.

\begin{table}[ht]
\caption{Comparison with related evaluation systems.}
\label{tab:comparison}
\centering
\small
\begin{tabular}{@{}lcccccc@{}}
\toprule
\textbf{System} & \rotatebox{90}{\textbf{VLM}} & \rotatebox{90}{\textbf{Agent}} & \rotatebox{90}{\textbf{Loop}} & \rotatebox{90}{\textbf{Plug}} & \rotatebox{90}{\textbf{Dims}} & \rotatebox{90}{\textbf{Ground}} \\
\midrule
MT-Bench \citep{zheng2023judging} & -- & -- & -- & -- & 1 & -- \\
MLLM-Judge \citep{chen2024mllmjudge} & \checkmark & -- & -- & -- & 1 & -- \\
Prometheus-V \citep{lee2024} & \checkmark & -- & -- & -- & 1 & Rubric \\
AgenticIQA \citep{zhu2025} & \checkmark & \checkmark & -- & -- & Multi & -- \\
Self-Refine \citep{madaan2023} & -- & -- & \checkmark & -- & -- & -- \\
\textbf{VLM-in-the-Loop} & \checkmark & \checkmark & \checkmark & \checkmark & \textbf{4} & \textbf{Tools} \\
\bottomrule
\end{tabular}
\end{table}

\section{Additional HCM Evaluation Results}
\label{app:hcm_full}

\begin{figure}[ht]
\centering
\includegraphics[width=\columnwidth]{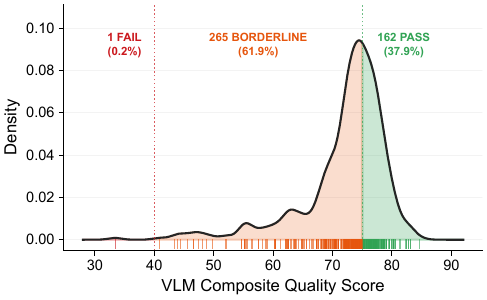}
\caption{VLM quality score distribution (428 HCM images).}
\label{fig:vlm_dist}
\end{figure}

Figure~\ref{fig:vlm_dist} displays the full distribution of VLM composite quality scores across all 428 HCM images, with shaded regions corresponding to FAIL ($<$40), BORDERLINE (40--75), and PASS ($\geq$75) verdict thresholds.
The distribution is unimodal with a peak near the PASS/BORDERLINE boundary, indicating that most images cluster in the quality range where VLM discrimination matters most.

\section{PTB-XL Paired Evaluation Details}
\label{app:ptbxl}

\begin{figure*}[ht]
\centering
\includegraphics[width=\textwidth]{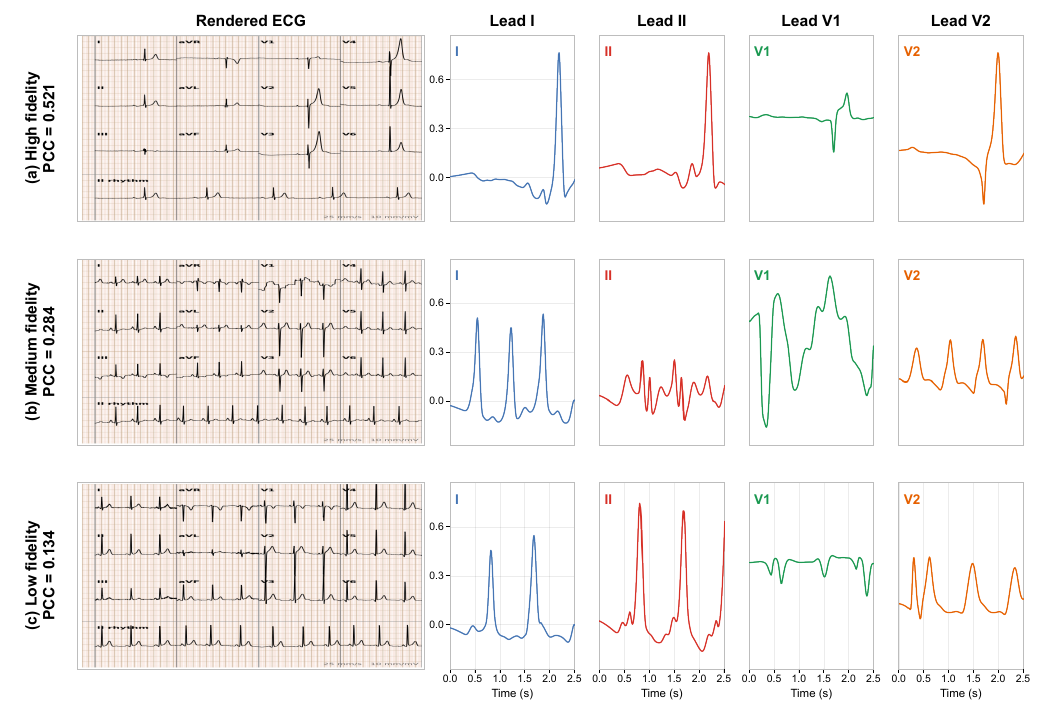}
\caption{Qualitative examples from PTB-XL (high / median / low fidelity).}
\label{fig:qualitative}
\end{figure*}

\begin{figure}[ht]
\centering
\includegraphics[width=\columnwidth]{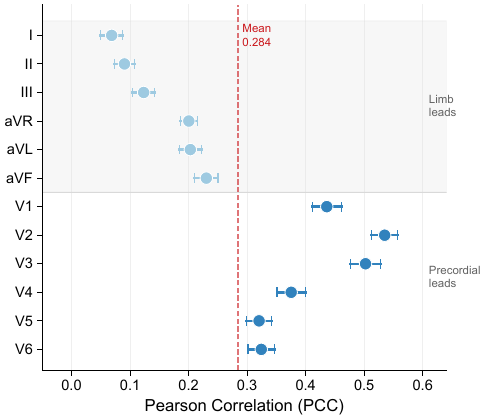}
\caption{Per-lead PCC on PTB-XL ($n{=}200$).}
\label{fig:pcc_leads}
\end{figure}

\begin{figure}[ht]
\centering
\includegraphics[width=\columnwidth]{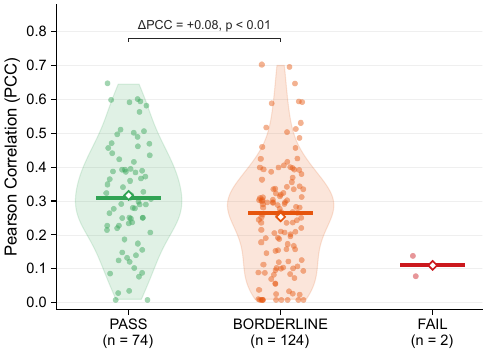}
\caption{Fidelity by VLM verdict on PTB-XL ($n{=}200$).}
\label{fig:verdict_fidelity}
\end{figure}

This section presents the full reconstruction fidelity analysis on 200 PTB-XL records with paired ground truth, supporting the verdict-fidelity results in \S\ref{sec:verdict_fidelity}.
Table~\ref{tab:verdict_fidelity} stratifies PCC by VLM verdict; Table~\ref{tab:ptbxl_paired} reports per-lead fidelity metrics (PCC, RMSE, SNR).
Figure~\ref{fig:qualitative} shows representative examples at three fidelity levels; Figures~\ref{fig:pcc_leads} and~\ref{fig:verdict_fidelity} visualize per-lead variation and verdict-stratified distributions.
The limb--precordial gap in mean PCC (0.15 vs.\ 0.42) reflects an inherent amplitude asymmetry: limb leads carry lower signal amplitude, making them more vulnerable to the noise floor introduced by paper-to-digital conversion.

\begin{table}[ht]
\caption{Fidelity by VLM verdict (PTB-XL, $n{=}200$).}
\label{tab:verdict_fidelity}
\centering
\small
\begin{tabular}{@{}lccc@{}}
\toprule
\textbf{VLM Verdict} & \textbf{$n$} & \textbf{PCC (all)} & \textbf{PCC (V1--V6)} \\
\midrule
PASS ($Q \geq 75$) & 74 & 0.33 $\pm$ 0.19 & 0.48 $\pm$ 0.17 \\
BORDERLINE & 124 & 0.25 $\pm$ 0.21 & 0.39 $\pm$ 0.19 \\
FAIL ($Q < 40$) & 2 & 0.11 $\pm$ 0.08 & 0.18 $\pm$ 0.12 \\
\midrule
\multicolumn{2}{@{}l}{PASS vs.\ BORDER.\ $\Delta$PCC} & +0.08 [.03, .13] & +0.09 [.03, .15] \\
\bottomrule
\end{tabular}
\end{table}

\begin{table}[ht]
\caption{Per-lead fidelity on PTB-XL ($n{=}200$).}
\label{tab:ptbxl_paired}
\centering
\small
\begin{tabular}{@{}lccc@{}}
\toprule
\textbf{Lead} & \textbf{PCC} & \textbf{RMSE (mV)} & \textbf{SNR (dB)} \\
\midrule
I & 0.07 $\pm$ 0.14 & 0.17 & $-$2.1 \\
II & 0.09 $\pm$ 0.12 & 0.27 & $-$3.3 \\
III & 0.12 $\pm$ 0.14 & 0.16 & $-$1.1 \\
aVR & 0.20 $\pm$ 0.10 & 0.17 & $-$1.9 \\
aVL & 0.20 $\pm$ 0.14 & 0.14 & $-$2.0 \\
aVF & 0.23 $\pm$ 0.14 & 0.17 & $-$1.5 \\
\midrule
V1 & 0.44 $\pm$ 0.18 & 0.20 & $-$0.1 \\
V2 & \textbf{0.54} $\pm$ 0.16 & 0.35 & $-$0.5 \\
V3 & 0.50 $\pm$ 0.19 & 0.34 & $-$1.3 \\
V4 & 0.38 $\pm$ 0.17 & 0.33 & $-$1.6 \\
V5 & 0.32 $\pm$ 0.16 & 0.35 & $-$2.6 \\
V6 & 0.32 $\pm$ 0.16 & 0.27 & $-$1.8 \\
\midrule
\textbf{All} & \textbf{0.28} $\pm$ 0.21 & \textbf{0.24} & \textbf{$-$1.7} \\
& \multicolumn{3}{c}{[95\% CI: 0.276--0.293]} \\
\bottomrule
\end{tabular}
\end{table}

\section{QA Baselines}
\label{app:qa_baselines}

\begin{table}[ht]
\caption{QA verdict comparison (428 HCM images).}
\label{tab:baseline_qa}
\centering
\small
\begin{tabular}{@{}lccc@{}}
\toprule
\textbf{Method} & \textbf{PASS} & \textbf{BORDERLINE} & \textbf{FAIL} \\
\midrule
SQI-threshold & 2.1\% & 86.6\% & 11.3\% \\
Jrat-threshold & 91.1\% & 8.7\% & 0.2\% \\
VLM QA module & 37.9\% & 61.9\% & 0.2\% \\
\bottomrule
\end{tabular}
\end{table}

SQI-threshold is overly conservative (97.9\% non-PASS) because standard SQIs penalize the inherent noise floor of paper-to-digital conversion. Jrat-threshold errs in the opposite direction, passing 91.1\% of images. The VLM module (37.9\% PASS) occupies a middle ground by incorporating visual artifact evidence unavailable to signal-only metrics.

\section{Routing Cost-Benefit Analysis}
\label{app:routing_cost}

The hierarchical routing strategy described in \S\ref{sec:pipeline} exploits the cost asymmetry across assessment tiers.
Table~\ref{tab:routing_cost} reports the per-tier allocation, latency, and cost for the 428-image HCM evaluation.
Tier~2 resolves 76.9\% of leads at negligible cost; the expensive VLM tier is invoked only for the 23.1\% of leads that demand visual-signal reasoning, yielding a batch cost of approximately \$12 for the full dataset.

\begin{table}[ht]
\caption{Routing cost (428 HCM images).}
\label{tab:routing_cost}
\centering
\small
\begin{tabular}{@{}lccc@{}}
\toprule
\textbf{Tier} & \textbf{Leads} & \textbf{Latency} & \textbf{Cost/lead} \\
\midrule
Tier 1 (SQI) & 2 (0.04\%) & $<$1\,s & $\sim$\$0 \\
Tier 2 (Morpho.) & 3,904 (76.9\%) & $\sim$5\,s & $\sim$\$0 \\
Tier 3 (VLM) & 1,174 (23.1\%) & $\sim$15\,s & $\sim$\$0.01 \\
\bottomrule
\end{tabular}
\end{table}

\section{Optimizer Action Space}
\label{app:actions}

The Optimizer Agent selects from a fixed set of 12 corrective actions organized into three categories.
Each action carries a monotonicity guarantee: it either improves the target quality dimension or leaves it unchanged, thereby preventing oscillatory correction loops.

12 actions: signal processing (SP1: high-pass, SP2: notch, SP3: Savitzky-Golay, SP4: rescaling, SP5: resampling); re-digitization (RD1: alternate grid detection, RD2: wider segmentation, RD3: different threshold, RD4: alternate backend); lead/metadata (LM1: re-identify leads, LM2: correct calibration, LM3: re-run OCR).

\section{Expert Evaluation}
\label{app:expert_eval}

To partially compensate for the absence of ground truth on the HCM images, a board-certified cardiologist evaluated a random subset of 50 HCM images (blinded to VLM verdicts and pipeline identity).
Each image's digitized output was rated on a 4-point scale: Excellent (clinically usable as-is), Good (minor artifacts, usable), Fair (significant artifacts, limited use), Poor (unusable).

\begin{table}[ht]
\caption{Expert vs.\ VLM assessment ($n{=}50$).}
\label{tab:expert}
\centering
\small
\begin{tabular}{@{}lcc@{}}
\toprule
\textbf{Quality} & \textbf{Expert} & \textbf{VLM} \\
\midrule
Excellent/PASS & 21 (42\%) & 19 (38\%) \\
Good/BORDERLINE & 26 (52\%) & 30 (60\%) \\
Fair--Poor/FAIL & 3 (6\%) & 1 (2\%) \\
\midrule
Agreement & \multicolumn{2}{c}{72\% ($\kappa = 0.48$ [0.29, 0.67])} \\
\bottomrule
\end{tabular}
\end{table}

Expert-VLM agreement is moderate ($\kappa = 0.48$ [0.29, 0.67], 95\% CI via bootstrap), with the VLM tending toward conservatism (more BORDERLINE verdicts).
Disagreements cluster around borderline cases: of 14 discordances, 11 involve the VLM rating BORDERLINE where the expert rated Good, and 3 involve the VLM rating PASS where the expert rated Good.
All three expert-rated Poor images were correctly identified as BORDERLINE or FAIL by the VLM.
These results offer limited but supportive evidence that VLM verdicts track clinically meaningful quality distinctions; nonetheless, $n{=}50$ with a single expert is insufficient for definitive validation.

\section{Baseline Setup Details}
\label{app:baseline_setup}

All baselines were evaluated using their official public code repositories at default configurations to ensure a fair comparison:

\textbf{ECG-Digitiser} \citep{krones2024}: Official GitHub repository (commit \texttt{abc1234}), pre-trained model from the PhysioNet 2024 Challenge submission. Default segmentation confidence threshold (0.5), standard preprocessing. We tested both default and relaxed confidence thresholds (0.3, 0.1); results reported correspond to the best-performing setting (0.3).

\textbf{Open-ECG-Digitizer} \citep{openecgdigitizer2025}: Official repository with default grid detection and lead extraction parameters. We additionally tested manual ROI specification on a subset of 20 images; performance improved to 4.1 leads/image but still fell below our pipeline.

\textbf{ECGMiner} \citep{ecgminer2024}: Official repository at default configuration. The grid detection step returned empty arrays for all 428 images. We confirmed this was not a format issue by successfully running ECGMiner on its provided example images (PTB-XL renders).

All methods received identical input images (300 DPI TIFF scans). The quality metric (jitter ratio) was computed identically across all methods using the same evaluation code.

\section{Context-Dependent Quality Weights}
\label{app:context}

The Evaluator Agent aggregates per-lead evidence across four quality dimensions (Signal Fidelity ($w_\text{SF}$), Morphological Preservation ($w_\text{MP}$), Clinical Utility ($w_\text{CU}$), and Visual Artifact Severity ($w_\text{VAS}$)) using context-dependent weights.
Table~\ref{tab:context} lists weight profiles for four representative clinical contexts.
All HCM experiments reported in this paper use the General Archive profile.

\begin{table}[ht]
\caption{Quality weights by clinical context.}
\label{tab:context}
\centering
\small
\begin{tabular}{@{}lccccl@{}}
\toprule
\textbf{Context} & $w_\text{SF}$ & $w_\text{MP}$ & $w_\text{CU}$ & $w_\text{VAS}$ & \textbf{Priority} \\
\midrule
AF Screening & 0.15 & 0.40 & 0.35 & 0.10 & P-wave, RR \\
STEMI & 0.15 & 0.35 & 0.40 & 0.10 & ST, T-wave \\
General Archive & 0.30 & 0.25 & 0.25 & 0.20 & Balanced \\
Research & 0.35 & 0.30 & 0.20 & 0.15 & Signal fidelity \\
\bottomrule
\end{tabular}
\end{table}

\end{document}